\documentclass{article}

\usepackage{arxiv}

\usepackage[utf8]{inputenc} % allow utf-8 input
\usepackage[T1]{fontenc}    % use 8-bit T1 fonts
\usepackage{hyperref}       % hyperlinks
\usepackage{url}            % simple URL typesetting
\usepackage{booktabs}       % professional-quality tables
\usepackage{amsfonts}       % blackboard math symbols
\usepackage{nicefrac}       % compact symbols for 1/2, etc.
\usepackage{microtype}      % microtypography

\usepackage{graphicx}
\usepackage{xcolor}
\usepackage{amssymb}
\usepackage{mathptmx}           % use Times fonts if available on your TeX system
\usepackage[mathscr]{euscript}  % use this package to get the curvy-L symbol for loss function
\usepackage{multirow}

\renewcommand{\headeright}{}
\renewcommand{\undertitle}{}

\title{Vector Embeddings with Subvector Permutation Invariance using a Triplet Enhanced Autoencoder}
\author{
  Mark~Alan~Matties\\
  The~Johns~Hopkins~University~Applied~Physics~Laboratory\\
  Laurel, MD 20723\\
  \texttt{mark.matties@jhuapl.edu}\\
}

\begin{document}
\maketitle

\begin{center}
\textit{DISTRIBUTION STATEMENT A: Approved for public release; distribution is unlimited.}\\
\end{center}
\vspace{2em}

\begin{abstract}
The use of deep neural network (DNN) autoencoders (AEs) has recently exploded due to their wide applicability. However, the embedding representation produced by a standard DNN AE that is trained to minimize only the reconstruction error does not always reveal more subtle patterns in the data. Sometimes, the autoencoder needs further direction in the form of one or more additional loss functions. In this paper, we use an autoencoder enhanced with triplet loss to promote the clustering of vectors that are related through permutations of constituent subvectors. With this approach, we can create an embedding of the vector that is nearly invariant to such permutations. We can then use these invariant embeddings as inputs to other problems, like classification and clustering, and improve detection accuracy in those problems.
\end{abstract}

% keywords can be removed

\section{Introduction}
\label{sec:intro}
For most problems solved using neural networks, the features (ie, elements of input vectors), are distinguishable. One example is weather related data, which might have features like temperature, wind speed and humidity. While these quantities could occur in any order in the input vector, the neural network designer picks one particular order and the network is trained with data fed to it in that order. Patterns that occur in the data across inputs are important and should be captured in the DNN model. However, for some problems, the features may be \textit{internally indistinguishable},\textit{interchangeable} or \textit{functionally equivalent}. That is, their order really does not matter and should not be interpreted as important. In this case, a pattern like increasing or decreasing values across the inputs would actually be spurious. For example, a feature vector might represent a move in a game where one part of the feature vector represents units that are allocated across four corps (ie, groups). This assignment is  represented by a four element subvector\footnote{Throughout this work, each subvector has dimension 4.} of the feature vector. In this example, the corps are functionally equivalent -- there is no meaningful difference between them\footnote{Note that it is true for this example. The order might be meaningful in other examples.}. So, assigning units of sizes of 5, 6, 8, and 9 to the four groups as \mbox{\{5, 6, 8, 9\}} is actually no different than \mbox{\{8, 6, 9, 5\}}. A DNN that gives more importance to \mbox{\{5, 6, 8, 9\}} because it is monotonically increasing would have discovered what is, in this case, a meaningless pattern. Another example might be where a subvector holds information on a direction moved (eg, \mbox{\{up, down, left, right\}} or \mbox{\{N, S, E, W\}}), where, aside from the (arbitrary) labels, the directions are indistinguishable. We do not want the neural network to learn such spurious patterns and so endeavored to develop a neural network that creates equivalent, or invariant, representations of such internally indistinguishable inputs that are permutations of each other.

The work we relate here was motivated by an original problem of training a reinforcement learning agent to play a strategy game, the output of which we want to analyze for clusters of behavior relating patterns of play by the agent \textit{on a per turn basis} to game results (win, tie, lose), where the games vary in length. We could have simply used a classifier based on a long short-term memory (LSTM) neural network for this problem except for the fact that the subvectors  are internally indistinguishable. Again, we do not want the classifier to possibly identify monotonically increasing or decreasing (or other spurious) patterns across the four corps subvector as significant. Thus, we took a two stage approach. In the first stage, we sought to create an neural network that, for each turn vector, would provide a representation that was invariant to such permutation. Then, in the second stage, each of these turn representations would be fed into an LSTM classifier. This paper covers the work of the first stage and demonstrates that a triplet loss enhanced autoencoder neural network can produce permutation invariant embedding representations of the input vectors.

\begin{figure}[t] 
  \centering
  \includegraphics[height=3cm]{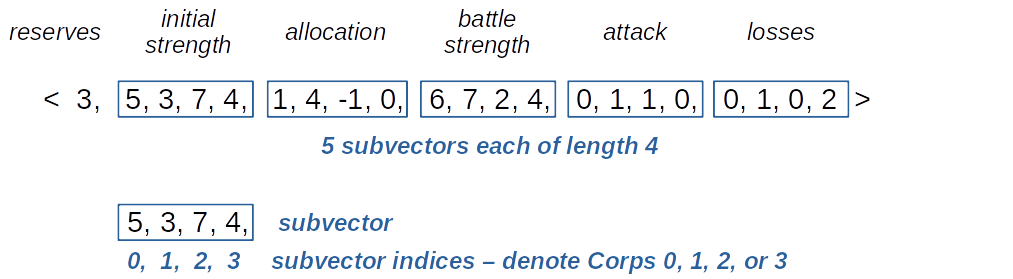}
  \caption{Example of a turn vector from the motivating game. The vector consists of a scalar at index 0 followed by five subvectors each of length 4.}
  %% NOTE!! \label{} must go AFTER the caption for the reference to the figure to work correctly
  \label{fig:game_vector_figure}
\end{figure}

In the motivating strategy game, the feature vector comes from turns for which the input vector contains several subvectors (representing different \textit{properties} within the vector) that are indistinguishable by order, but are linked. There are four corps ($A, B, C, D$), denoted by subvector indices \mbox{(0, 1, 2, 3)} in Figure \ref{fig:game_vector_figure}, and several properties ($P_{1}$, $P_{2}$, etc.). Any ordering of corps in the properties (ie, subvectors) is acceptable as long as it is consistent across all properties. To illustrate, the following two representations are completely equivalent and should be treated as such by a neural network -- \hfill \break
\centerline{$\langle \; P_{1}(A,B,C,D), \; P_{2}(A,B,C,D) \; \rangle \; \longleftrightarrow \; \langle \; P_{1}(B,A,D,C), \; P_{2}(B,A,D,C) \; \rangle$.} 
As mentioned above, we seek to classify a \textit{sequence} of such vectors, where the sequence represents a series of turns taken during the game. Example classes are game result (\{win, tie, lose\}) and style of play (\{aggressive, moderate, passive\}). However, we do not want the classifier to learn spurious patterns like those mentioned above. 

To simplify the development and testing of our approach, we created a proxy data set that is similar to that of the game turn sequences, but has better properties that simplify assessing the efficacy of the proposed method. The proxy data is described in detail below. Since an autoencoder is designed to reproduce the input as the output and its reconstruction error (ie, between input and produces output) is usually distance metric based (ie, mean squared error between input and expected output), we do not expect a standard autoencoder to suffice. As a very simple example, consider the following two sets of vector permutation pairs: \mbox{A = $\{a_{0}, a_{1}\}$ = $\{\langle 1, 9 \rangle, \langle 9, 1 \rangle\}$} and \mbox{B = $\{b_{0}, b_{1}\}$ = $\{\langle 2, 8 \rangle, \langle 8, 2 \rangle\}$}. Here, $a_{0}$ and $a_{1}$ are permutations of each other, as are $b_{0}$ and $b_{1}$. We would like $a_{0}$ and $a_{1}$ to be in the same cluster as calculated by a distance metric, and $b_{0}$ and $b_{1}$ to be in their own cluster that is clearly separated from the $A$ cluster. However, feeding data like this into a standard autoencoder will result in embeddings $E(*)$ that place $E(a_{0})$, $E(b_{0})$ closer together and $E(a_{1})$, $E(b_{1})$ in another due to the euclidean proximity of their original input values. Clearly, we need to supplement the standard autoencoder with an additional constraint (in the form of a loss function) to learn the underlying structure that we want (clustering by permutation), preferably without introducing too much additional error or structure not related to the permutation relationship. Toward this end, we present work that supplements the standard reconstruction error of an autoencoder (mean squared error loss) with the triplet loss function.

% = new section = = = = = = = = = = = = = = = = = = = = = = = = = = = = = = = = = = = = 
\section{Background and Related Work}
\label{sec:related_work}

% - new subsection - - - - - - - - - - - - - - - - - - - - - - - - - - - - - - - - - - -
\subsection{The Autoencoder}
\label{sec:related_work::subsec:autoencoders}
An \textit{autoencoder} is a type of neural network that is specially designed to reconstruct its input vector at the output while generating an intermediate embedding representation of the original input \cite{Goodfellow2016Chapter14}. It consists of two internal subnetworks that are often structural mirror images of each other (the \textit{encoder} and \textit{decoder}). These subnetworks can have several layers and the autoencoder itself can be created using different neural network architectures, such as a multilayer perceptron (MLP), a long short-term memory recurrent neural network (LSTM/RNN), or a convolutional neural network (CNN). The output from the last layer of the encoder subnetwork (that feeds the input of the decoder subnetwork) provides an intermediate representation of the input data and is referred to as the \textit{embedding representation} or, simply, the \textit{embedding}. An autoencoder is typically trained to reproduce its inputs with high accuracy by minimizing the error in reconstructing the value of the input at the output. This error is the \textit{reconstruction error}. The mean squared error loss function (\textit{mse loss}) is often used to track reconstruction error during training. By reducing this error to very low levels, the embedding becomes a high fidelity proxy for the original input vector. 

The embedding dimension may be less than (\textit{undercomplete}), greater than (\textit{overcomplete}), or the same as the original input vector dimension. A popular use of the undercomplete autoencoder is for \textit{dimensionality reduction}, where the embedding dimension is much less than that of the original vector. Overcomplete autoencoders are often used in denoising applications. A general review of autoencoders and their applications can be found in \cite{TschannenAutoencoderAdvances_Proceedings}.

% - new subsection - - - - - - - - - - - - - - - - - - - - - - - - - - - - - - - - - - -
\subsection{Triplet Loss}
\label{sec:related_work::subsec:triplet_loss}
Triplet loss was introduced by Chechik et al. \cite{chechik2010large} and popularized by Schroff et al. \cite{schroff_facenet_2015}. Using triplet loss, the training goal is to reduce the ``distance'' between two samples that are similar, eg, they belong to the same class, and to increase the distance between two samples that are dissimilar, eg, they belong to different classes. The ``distance'' is usually a \textit{distance metric}, such as the euclidean distance. So, triplet loss belongs to the class of problems called deep metric learning \cite{hoffer_deep_2015_proceedings}, \cite{ishfaq_deep_nodate}, \cite{kaya_deep_2019}, \cite{KumarSemiSupervisedClusteringWithMetricLearning2005}.

\begin{figure}[t] 
  \centering
  \includegraphics[height=5cm]{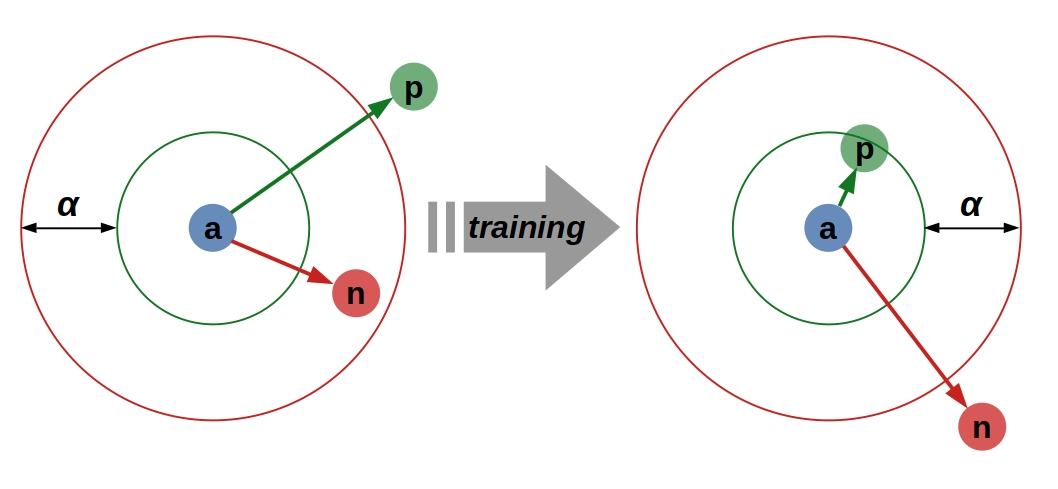}
  \caption{Illustration of the triplet loss constraint showing three embedding samples - anchor (blue \textit{a}), positive (green \textit{p}) and negative (red \textit{n}) with their distance relation before and after training. The anchor and positive belong to the same class, while the anchor and negative do not. Thus, the positive and negative also do not belong to the same class. In this example, the anchor and negative are closer than the anchor and positive. Through training, the negative is pushed farther away and the positive is drawn closer to the anchor to an extent driven by the triplet loss margin $\alpha$, thus reducing the error represented by Equation \protect\ref{eqn:triplet_loss}.}
  %% NOTE!! \label{} must go AFTER the caption for the reference to the figure to work correctly
  \label{fig:triplet_loss_figure}
\end{figure}

As shown in Figure \ref{fig:triplet_loss_figure}, the \textit{anchor} point is a sample to which two other samples with known relationship to the anchor will be compared. The \textit{positive} point has some similarity to the anchor while the \textit{negative} point is dissimilar to the anchor. In our work, similar points belong to the same permutation point set and dissimilar points belong to different point sets. The degree of dissimilarity can be adjusted using the parameter $\alpha$, the \textit{triplet loss margin}. This parameter drives the magnitude of the difference between anchor-positive (\textit{a-p}) pair distance and the anchor-negative \textit{a-n} pair distance. Note that not all \textit{a-p} and \textit{a-n} pairs will necessarily have the relationship shown in Figure \ref{eqn:triplet_loss}. There will be a mix, with some \textit{a-p} pairs already close and some \textit{a-n} pairs already far, perhaps already by the margin $\alpha$. Such sets of pairs, or \textit{a-p-n triplets} do not contribute much to the training of the neural network because the weights in the network do not change much or at all when presented these inputs. (They are called \textit{easy triplets}.) A goal is to choose triplets that have a relationship more like that shown in Figure \ref{fig:triplet_loss_figure}, where the weights of the neural network would have to be substantially adjusted to achieve the desired post-training separation. Such triplets are called hard or semi-hard. Triplet selection can add substantial computational time to training a neural network using triplet loss \cite{schroff_facenet_2015} and has become a focus of study itself, called \textit{triplet mining} \cite{schroff_facenet_2015}.

When used as a loss function in training a neural network, the triplet loss has the form given by Equation \ref{eqn:triplet_loss}, where $a, p, n$ are the anchor, positive, and negative samples, $f(*)$ is the embedding representation of the sample, and $\alpha$ is the triplet loss margin, the desired minimum separation between positive and negative pairs. The metric, given by $\| * \|$, is usually taken to be a euclidean distance, but in principle can be any appropriate metric.
\begin{equation}
  \textit{L}(a, p, n) = \textrm{max}(\|f(a)-f(p)\|^{2} - \|f(a)-f(n)\|^{2} + \alpha,\;\;0)
  \label{eqn:triplet_loss}
\end{equation}
The goal in training the neural network then is to minimize the sum of all triplet loss errors over all input samples by adjusting the weights of the network. As a result, some samples will be pulled closer together and others will be pushed farther apart, allowing for classification and clustering based on separation distance. In fact, triplet loss has enjoyed some success in classifying images \cite{schroff_facenet_2015}, \cite{dong2018triplet}, \cite{hermans_defense_2017}.

% - new subsection - - - - - - - - - - - - - - - - - - - - - - - - - - - - - - - - - - -
\subsection{Autoencoders using Triplet Loss}
\label{sec:related_work::subsec:autoencoders_and_triplet_loss}
Some work has combined autoencoders and triplet loss for classification, however, not as a means to generate invariant representations. In \cite{yang_triplet_2019}, the authors used a triplet loss enhanced autoencoder to encode graphs, specifically, social networks. They claim this approach preserves (social) network topology while capturing clustering structure. They created three different autoencoders to treat three different social networks each of which had different input dimensions and used an embedding dimension of 100 for all autoencoders. The encoder part had either two or three layers (for two data sets and the third data set, respectively) and they used the \textit{mahalanobis distance} for triplet loss metric. For one data set, they briefly looked at the effect of varying $\alpha$ over the values [0.01, 0.1, 1, 10, 100] on Macro-F1 for classification and Normalized Mutual Information (NMI) for clustering, and show that the results for this data set are mostly \textit{insensitive} to choice of $\alpha$. However, they do not report the effect on cluster structure, including whether any artifacts are introduced by the choice of $\alpha$.

In \cite{ienco_deep_triplet_2019}, researchers used a two stage process involving a triplet loss enhanced autoencoder to generate embeddings, followed by a refinement of the embedding values to enhance their clustering properties. For the second stage, they take the \textit{encoder subnetwork} of the autoencoder (trained in the first stage), then further train it to ``refine'' the clusters of the output embeddings. Since their objective ends with determining the clustering behavior, this approach may suffice. However, such subsequent training likely creates an uncontrolled divergence from the embedding values that best reproduce the original inputs. As such, this second step likely increases reconstruction error, in our opinion.\footnote{This effect is neither discussed nor evaluated in their paper.} In our work, maintaining the best reconstruction of the original inputs with the lowest error is important, so this approach is neither suitable nor, as it turns out, necessary.

% = new section = = = = = = = = = = = = = = = = = = = = = = = = = = = = = = = = = = = =
\section{Approach}
\label{sec:approach}
% - new subsection - - - - - - - - - - - - - - - - - - - - - - - - - - - - - - - - - - -
\subsection{Data Generation and Characterization}
\label{sec:approach::subsec:test_data}
\begin{figure}[t]
  \includegraphics[width=\textwidth]{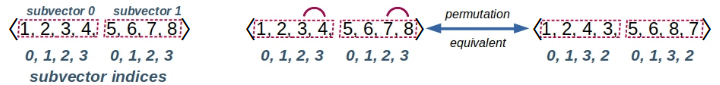}
  \caption{Illustration of subvector structure and its permutation. In this work, the permuted vectors are equivalent and should map to the same embedding vector.}
  \label{fig:subvector_figure}
\end{figure}

\paragraph{Terminology.} Not all sets of points may form clusters. We expect points that belong to the same cluster to be close in their native coordinate space and the clusters themselves to be well separated, preferably with no overlap. Such clusters would then be \textit{discernible}, in that we can easily see that there is separation into well defined groups. In this work, we distinguish between a set of points related by consistent permutation across subvectors as \textit{permutation point sets} and whether the constituent points of those sets form clusters in space with clear separation as \textit{permutation clusters}, or simply \textit{clusters}.
As shown in Figure \ref{fig:subvector_figure}, consider the example vector \mbox{$\langle$1, 2, 3, 4, 5, 6, 7, 8$\rangle$}, which has two subvectors -- \mbox{$\langle$1, 2, 3, 4$\rangle$} and  \mbox{$\langle$5, 6, 7, 8$\rangle$}. We can permute this vector by swapping elements in the last two positions of every subvector, that is, 
\mbox{$\langle$1, 2, \textcolor{red}{3}, \textcolor{red}{4},  5, 6, \textcolor{red}{7}, \textcolor{red}{8}$\rangle$} becomes \mbox{$\langle$1, 2, \textcolor{red}{4}, \textcolor{red}{3},  5, 6, \textcolor{red}{8}, \textcolor{red}{7}$\rangle$}. If we do this for every possible unique permutation of subvector indices, we will have 24 vectors in one permutation point set, including the original one. Whether all vectors in this permutation point set also form a cluster in space is a different matter. As discussed above in the Introduction, we do not expect them to be clustered.

%% - - - - data generation
\paragraph{Data Generation.}
To test the ability of triplet loss to drive permutation clustering in an autoencoder, we generated a simplified, proxy data set consisting of vectors with dimension 24 and subvector dimension of 4, giving 6 subvectors per vector. We chose vector and subvector dimensions that reflected the original data in our problem. In the original data, each vector has dimension 25, but in structure it contains a scalar followed by a 24 dimensional vector with 4 dimensional subvectors (representing the 4 corps). Not including the scalar in this test data set is one of the simplifications we chose.

A 4 dimensional subvector produces 24 unique permutations for each vector with subvector indices ranging from \mbox{(0, 1, 2, 3)} to \mbox{(3, 2, 1, 0)}. The range of values of each element is the interval \mbox{[0, 24]} (not related to the number of subvector permutations) and is randomly chosen with replacement. A total of 40,000 original vectors were randomly generated and checked that none was a direct, complete permutation of any others in order to start with all unique points in the test data set. However, we allowed repeated values in a vector, and even a subvector, to remain. For each original data vector, its 23 unique permutations were generated to give a 24 vector permutation point set. This process yielded a proxy data set of 960,000 vectors total. 

To create training and analysis data sets, we removed 1,000 permutation point sets (or 24,000 total points) from the full data set and then reserved for post-training validation and analysis. For training the autoencoders, the train/test\footnote{We use the term \textit{test data} to denote data from the training set used to test the model during training and \textit{validation data} to mean data that was not seen during training and was used to evaluate the model post-training.} split of the remaining data set was 80/20, giving 31,200/7,800 train/test permutation point sets or 748,800/187,200 total vectors for each. It is these permutation point sets for which we want to generate embeddings that are very tightly clustered and thus create a representation that is permutation invariant. That is, any vector in the permutation point set maps to the \textit{same embedding value}. Since there will inevitably be noise in the mapping produced by the autoencoder, the embedding value will not be precisely equal for all vectors in the embedding point set, but only approximately so. Thus, the embedding will not be strictly invariant, but it will be invariant within some error $\epsilon$ that we can quantify. For this work, we use \textit{invariant} to mean \textit{invariant within some specifiable error}.
\begin{figure}[t]
  \centering
  \includegraphics[width=\textwidth]{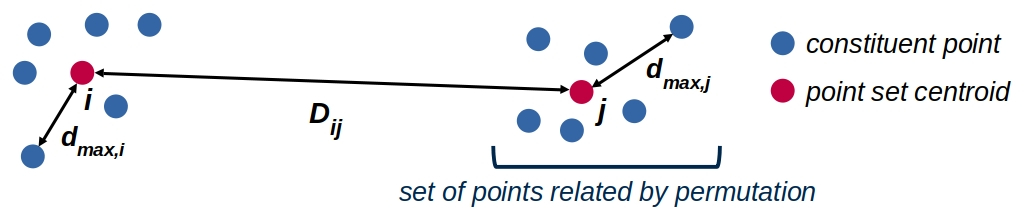}
  \caption{Illustration of permutation point sets, point set centroids, and their separation distances. These definitions apply both to points in the original vector space or the embedding space.}
  \label{fig:intra_inter_distances_figure}
\end{figure}

\paragraph{Data Characterization.}
To characterize the original data set, especially to verify that the permutation point sets did not form discernible clusters, we looked at two key measures - (1) the maximum separation distance of a point in the $i^{th}$ permutation point set from the point set centroid ($d_{max, i}$), and (2) the separation distance of centroids from two different point sets $i$ and $j$ ($D_{ij}$). These distances are illustrated in Figure \ref{fig:intra_inter_distances_figure}. Throughout this paper, we refer to $d_{max, i}$ as the \textit{max intra-point set distance} or the \textit{max intra-cluster distance}, and $D_{ij}$ as the \textit{inter-centroid distance}. If $D_{ij} >> (d_{max, i} + d_{max, j})$ for all permutation point set pairs $i, j$, then permutation point sets form well separated clusters. An upper bound on this inequality is $D_{ij} >> 2 \; d_{max}$ where $d_{max}$ is the largest value of $d_{max, i})$ for all $i$.

\begin{figure}[htb!]
  \centering
  \includegraphics[width=\textwidth]{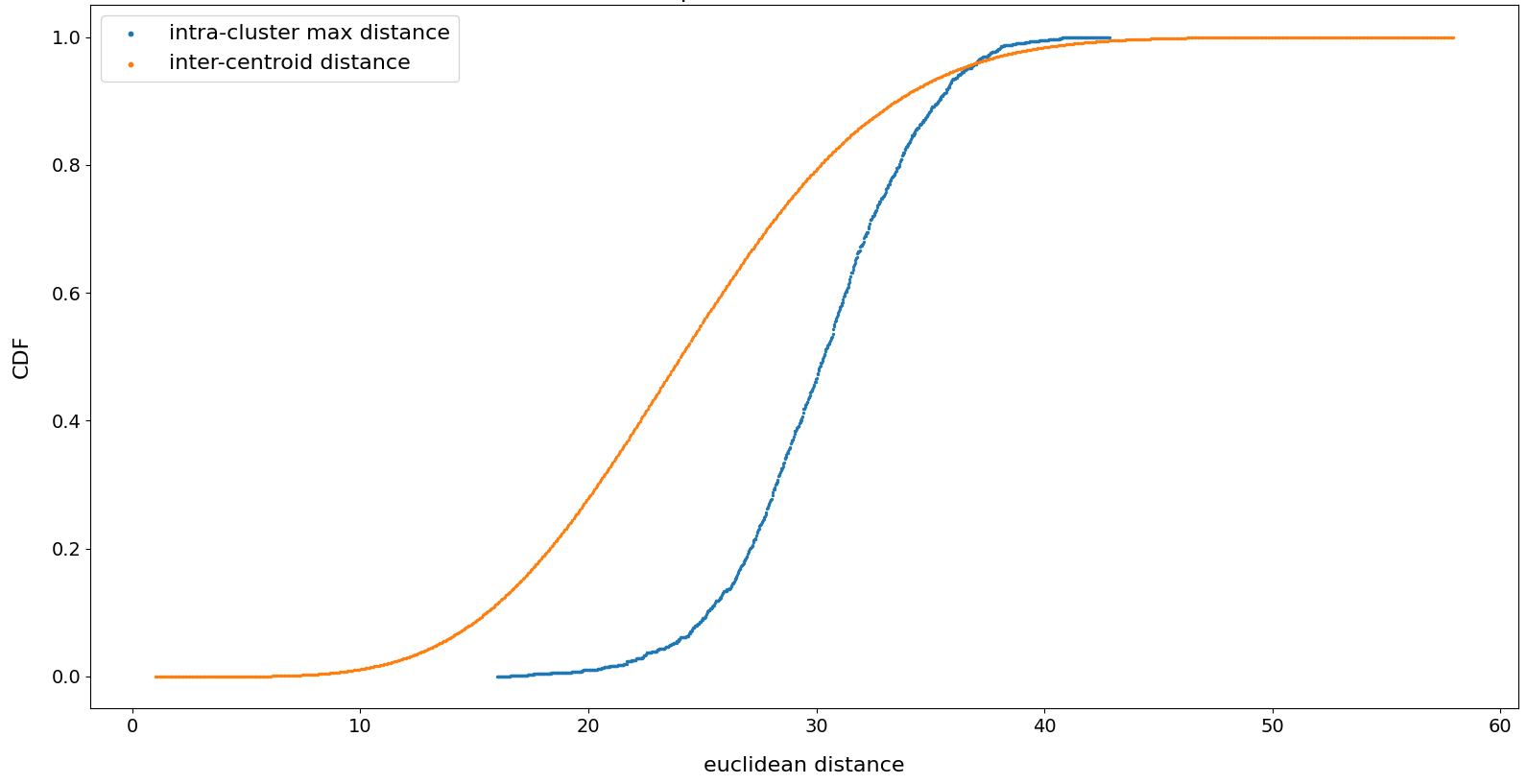}
  \caption{Cumulative distribution functions (CDFs) for (1) the maximum distance within a vector permutation point point set and the its centroid, $d_{max, i}$ (blue curve), and (2) the inter-centroid separation, $D_{ij}$ (red curve) for test set of 1000 original vectors (ie, not embeddings) and their permutations. Note that the separation of inter-centroid distances is less than or in the same range as the intra-point set maximum separation. This relationship shows that the dispersion of the potential clusters (ie, the intra-cluster maximum separation) is greater than or about the same as the separation of the clusters (ie, the inter-centroid separations). Thus, the ability to distinguish permutation based clusters using a distance metric \textit{in the original vector space} is virtually zero.}
  \label{fig:intra_and_inter_distance_for_vectors_and_centroids}
\end{figure}

To determine whether the permutation point sets naturally formed well separated clusters, we calculated the empirical cumulative distribution functions (CDF) of $d_{max, i}$ and $D_{ij}$ for the original data set and for the validation data set (1000 permutation point sets or 24,000 vectors), as shown in Figure \ref{fig:intra_and_inter_distance_for_vectors_and_centroids}. The key feature of this plot is that $d_{max, i}$ (\textit{intra-cluster max distance}, in blue) is greater than or in the same range as $D_{ij}$ (\textit{inter-centroid distance}, in red) for all points $i$ and all centroid pairs $ij$. That is, the maximum radii of any possible clusters was greater than the possible separation of those clusters. This relationship is the opposite of that illustrated in Figure \ref{fig:intra_inter_distances_figure}, but was what we expected in the original data - that there is significant overlap of permutation point sets in the original data and, thus, there cannot be any discernible clustering on this basis. Since the validation data set was randomly selected from the complete data set, we infer that the entire original data set (to be used for training the autoencoder) also does not exhibit clustering around permutation point sets. We then set out to construct an autoencoder that might yield embeddings of the original data that would exhibit the desired clustering. 

% - new subsection - - - - - - - - - - - - - - - - - - - - - - - - - - - - - - - - - - -
\subsection{Autoencoder Architecture}
\label{sec:approach::subsec:autoencoder_architecture}
\begin{figure}[htb!]
\begin{minipage}{0.55\textwidth}
  \centering
  \includegraphics[height=8cm]{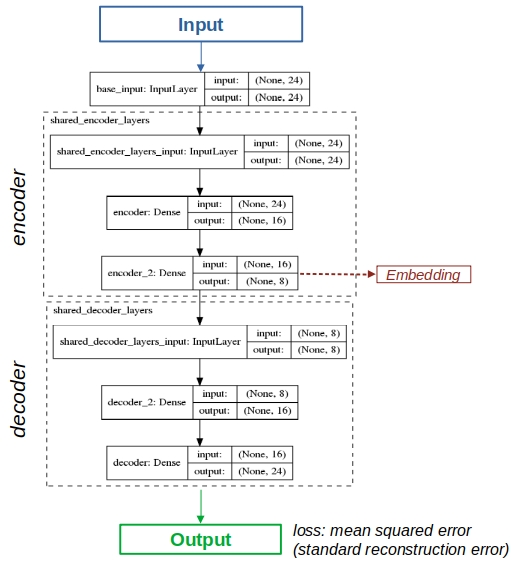}
%  \caption{.}
%  \label{fig:standard_autoencoder_network}
\end{minipage}
\begin{minipage}{0.45\textwidth}
  \centering
  \includegraphics[height=4.5cm]{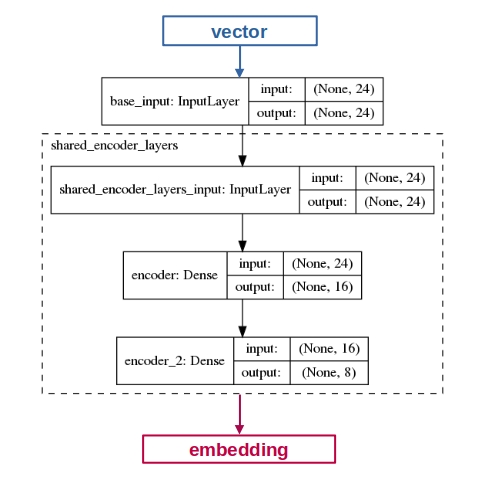}
%  \caption{encoder network.}
%  \label{fig:encoder}
\end{minipage}
  \caption{Standard autoencoder network (left) with the constituent encoder network (right).}
  \label{fig:standard_autoencoder_network}
\end{figure}

\begin{figure}[htb!]
  \centering
  \includegraphics[height=7.5cm]{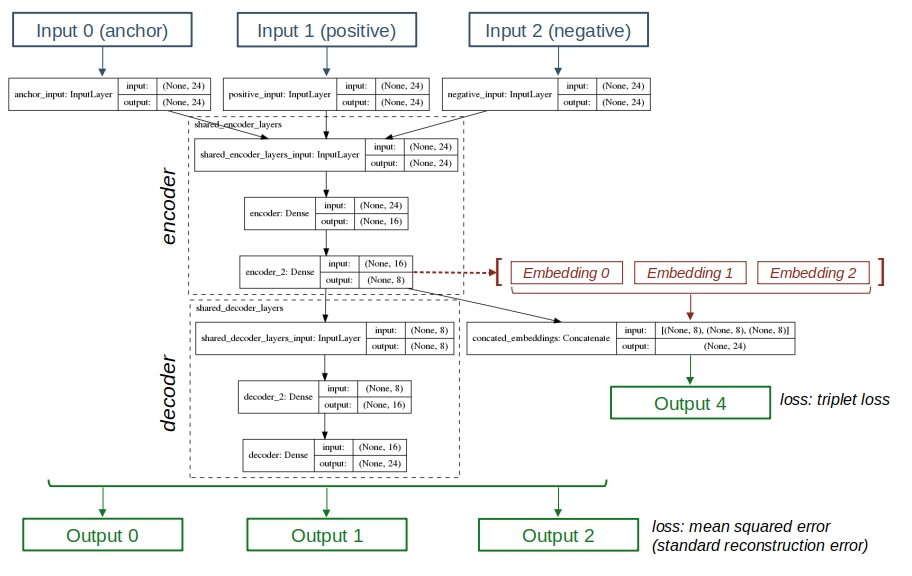}
  \caption{Enhanced autoencoder network with triplet loss.}
  \label{fig:enhanced_autoencoder_network}
\end{figure}

In order to generate embedding representations of the original vectors, we used two base autoencoder architectures - one with the usual mean squared error (\textit{mse}) loss function to minimize reconstruction error of the outputs and an enhanced version with an additional loss function, the \textit{triplet loss}. These autoencoders are shown in Figures \ref{fig:standard_autoencoder_network} and \ref{fig:enhanced_autoencoder_network} and we will refer to them as the \textit{standard autoencoder} and \textit{triplet loss enhanced autoencoder}, or simply \textit{enhanced autoencoder}, respectively. We created the enhanced autoencoder because the standard version did not yield the desired embeddings, as we discuss below. All neural networks and data analysis methods were implemented in Python 3.8.0 \cite{python3} using Keras 2.3.1 \cite{chollet2015keras}, Tensorflow 2.2.0 \cite{tensorflow2015-whitepaper}, NumPy 1.18.2 \cite{numpy}, and Scikit-Learn 0.22.2 \cite{scikit-learn}.

The standard autoencoder we used consists of a pair of stacked multilayer perceptrons (MLPs) (the \textit{encoder} and \textit{decoder}) that encodes the input into an embedding representation, then decodes it (ideally) back into its original value. We use an encoder that is two layers deep with each layer having an input and output size of 24 $\rightarrow$ 16 and 16 $\rightarrow$ 8, giving an embedding dimension of 8. The decoder layers invert this pattern with input and output sizes of 8 $\rightarrow$ 16 and 16 $\rightarrow$ 24. All layers use the \texttt{tanh} activation function except the final decoder layer, which uses a \texttt{sigmoid} function. The element values of the original vectors were scaled on [0, 1] for input into the autoencoder, so the sigmoid function, whose output is also on [0, 1], can reconstruct the scaled vectors. Input vectors were randomly shuffled. We used the Adam optimization function with a learning rate of 0.001 and the mean squared error loss function to minimize reconstruction error. Various other options were tested - more or fewer layers, number of nodes per layer, different activation functions, etc., though not exhaustively. The combination described here gave the best empirically observed performance for the problem at hand. \footnote{If a production quality solution were desired, we would run a systematic grid search to find the best combination of hyperparameters.}

The enhanced autoencoder starts with the same network architecture as the standard one, but incorporates an additional loss function -- the \textit{triplet loss} (as shown in Figure \ref{fig:enhanced_autoencoder_network}). In order to accommodate triplet loss, the enhanced autoencoder requires three inputs -- the anchor, positive, and negative inputs. It then provides three corresponding outputs. The reconstruction error (\textit{mse loss}) is evaluated for each corresponding input/output pair using shared network weights. Here, a positive sample is one from the same permutation point set as the anchor and the negative sample is from a different point set. The details on choosing triplets are discussed below.

Note that since the loss function is evaluated for each of the three input/output pairs, there are three times as many calculations as compared to the standard autoencoder. The triplet loss is calculated on the three embedding values and added to the mse loss to make up the total loss. The version of Keras/Tensorflow that we used does not provide a triplet loss function, so we wrote a custom one. During training, the weights in the network are adjusted to minimize the total loss. Keras/Tensorflow allows weighting of these two component losses. In this work, we used equal weighting of mse loss and triplet loss.

\subsection{Training the Autoencoder Models}
\label{sec:approach::subsec:training_autoencoder_models} 
We created and trained four different autoencoders -- a single standard one and three enhanced ones, each of which used a value for $\alpha$ of 1.0, 2.0, or 5.0. While $\alpha$ = 1.0 is often cited as a typical value, we wanted to gauge the effect of $\alpha$ on the training and the results. Given the relatively large training set of 936,000 samples (vectors), we used a batch size of 5000 and ran the training for 1200 steps. Note that since the number of inputs for the enhanced autoencoder is three, the effective sample size is 3 $\times$ 936,000 = 2,808,000, with some vectors appearing multiple times in the training set of the enhanced autoencoders as positive or negative samples. As mentioned above, the train/test split was 80/20. These values gave good convergence of model weights in an acceptable period of training time. We trained our models on a computer with a multicore 3GHz CPU and 64GB of RAM. On this machine, the standard model took about 35 minutes to train and used about 3 CPU cores on average while the enhanced models each took about 12 hours to train and used about 4 CPU cores on average. Clearly, the enhanced model comes at a much greater cost in training time. However, the results it produces are well worth it, as we show below.

 During training, we used the mse loss (ie, reconstruction error) for all autoencoders, and added triplet loss for enhanced autoencoders, to drive adjustment of the network weights. In regression problems, these loss functions are also often tracked as \textit{metrics} when using Keras/Tensorflow. However, these loss functions are absolute measures, not relative ones, and so they do not give a complete picture. Is an mse loss of 0.01 good or bad? It depends on the magnitude (value) of the the true vector that the network output is meant to approximate. If the true vector value is 1.0, the 1\% error is probably not bad. If the true vector value is 0.02, a 50\% error is probably not good. We used a custom \textit{numeric accuracy} metric that is a relative measure of the error to track the input vector reconstruction. It is a variation on the sum of squared errors, but approaches 1 as the error goes to zero, and is given by
\begin{equation}
    numeric\;accuracy = max(1 - \frac{\sum_{k}{(y_{pred, k} - y_{true, k})^{2}}}{\sum_{k}{(y_{true, k})^{2}}},\;\;0)
\end{equation}
where $y_{true}$ is the true value of the output (ie, the original vector), $y_{pred}$ is the predicted value of the vector (ie, the actual output), and $k$ is the vector element index. We constrain the range to $[0, 1]$ using the $max$ function, since negative values are no more informative than a value of $0$, which is meant to indicate no accuracy at all. Note that in Keras/Tensorflow, the \textit{metric} function is informational only -- it does not drive the adjustment of the network weights. Only the \textit{loss} function drives the adjustment.

\begin{figure}[t]
  \includegraphics[width=\textwidth]{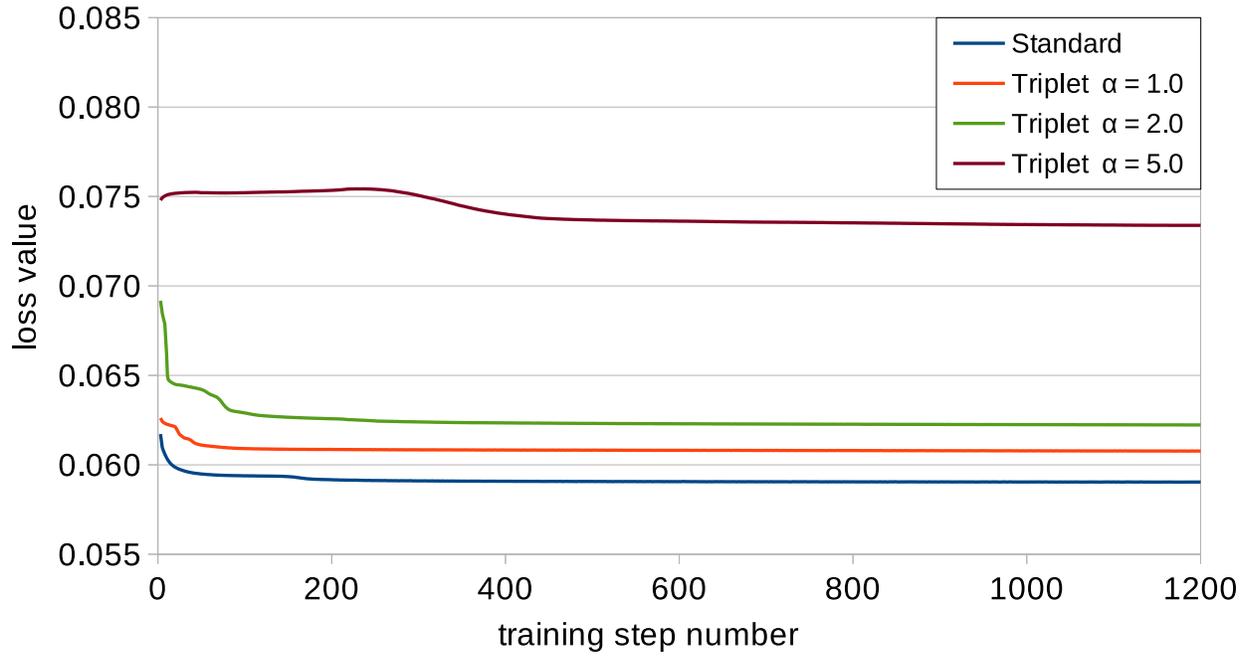}
  \caption{Reconstruction error (as mse loss) during training for each model -- the standard model and each of three enhanced models trained with $\alpha$ = 1.0, 2.0, and 5.0.}
  \label{fig:training_mse_loss}
\end{figure}

\begin{figure}[htb!]
  \includegraphics[width=\textwidth]{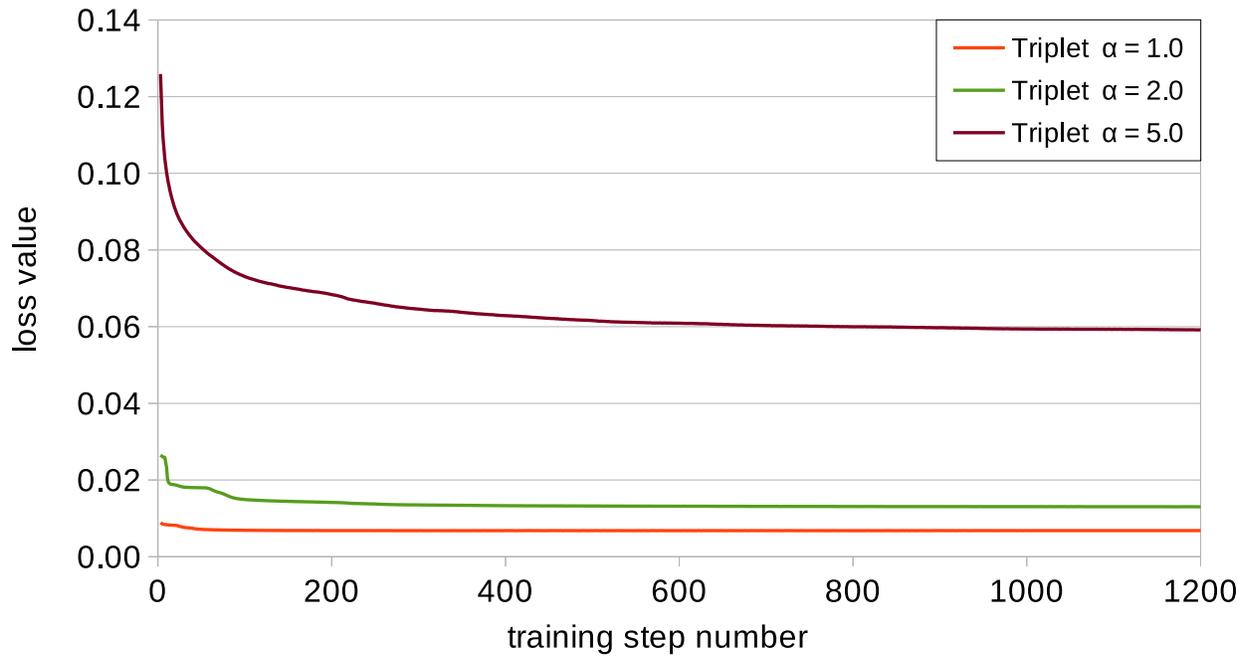}
  \caption{Triplet loss error during training for each of three enhanced models trained with $\alpha$ = 1.0, 2.0, and 5.0.}
  \label{fig:training_triplet_loss}
\end{figure}

\begin{figure}[htb!] 
  \includegraphics[width=\textwidth]{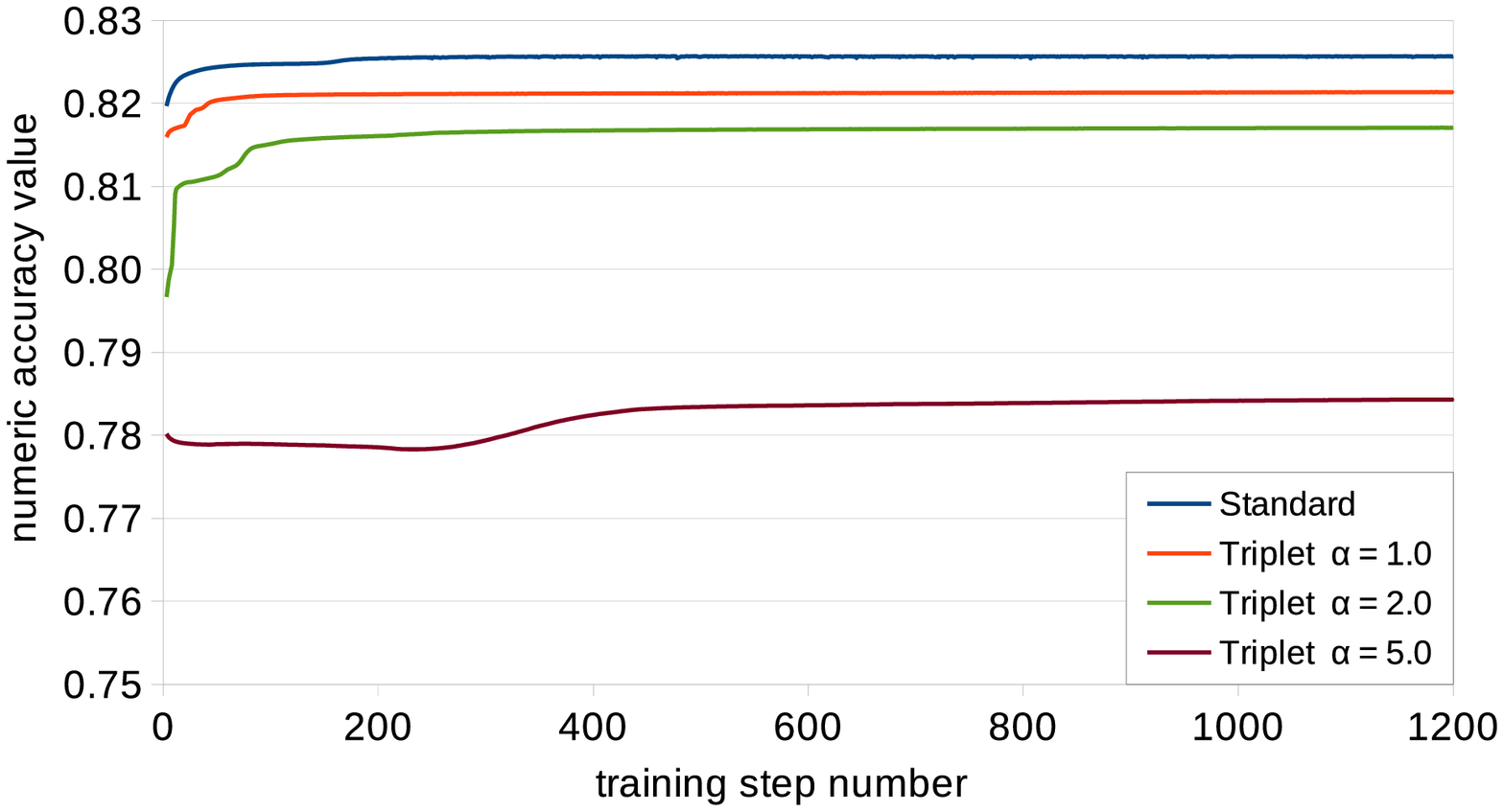}
  \caption{Numeric accuracy metric during training for each model -- the standard model and each of three enhanced models trained with $\alpha$ = 1.0, 2.0, and 5.0.}
  \label{fig:training_numeric_accuracy}
\end{figure}

We can see the effects of including triplet loss on model training in Figures \ref{fig:training_mse_loss}, \ref{fig:training_triplet_loss}, and \ref{fig:training_numeric_accuracy}. In Figure \ref{fig:training_mse_loss}, we see that there is an increase in the vector reconstruction error when we train the enhanced models vs. the standard model. Introducing triplet loss as a forcing function for permutation point set clustering comes at a cost that monotonically increases with the values of triplet loss margin $\alpha$ studied here, but faster than linearly. 

The additional triplet loss error also increases monotonically with the value of $\alpha$, as shown in Figure \ref{fig:training_triplet_loss}, and also superlinearly. A greater value of the triplet loss function implies that the embeddings produced by the enhanced autoencoder using that value of $\alpha$ is less able to separate positive and negative samples relative to the anchor by the minimum value of $\alpha$. However, we cannot assume that a higher triplet loss value for a greater $\alpha$, eg, $\alpha = 5.0$ vs. $\alpha = 1.0$, necessarily means that the former autoencoder does not separate embeddings as well as the latter, since the distance that the former tries to separate the embedding points is much greater than the latter. We present evidence of this fact below.

In training a neural network, the loss functions drive the adjustment of the weights, but the values of the losses (shown above) are not necessarily the best measure of the accuracy of the output produced by the model. In Figure \ref{fig:training_numeric_accuracy}, we see that the numeric accuracy of the outputs, ie, the accuracy of reconstructing the original input values, does decrease for the enhanced autoencoders vs. the standard one, and as $\alpha$ increases among the enhanced autoencoders, it does so monotonically and non-linearly. Thus, there is a cost in accuracy of output for the autoencoder when triplet loss is included and it is much worse for higher values of triplet loss margin $\alpha$, whether or not there are other potential benefits.

\paragraph{Choosing Triplets for Training.}

To minimize overtraining of the enhanced network, triplets were chosen so that every vector in the training set appeared as an anchor sample \textit{once}. For each anchor sample, a positive sample was randomly chosen from the same permutation point set. A negative sample was chosen by randomly selecting a different permutation point set, then randomly taking a point from it. We did not filter pairs on based on how close the negative sample was to the anchor in space. Thus, anchor/negative pair distances may have been less than or greater than anchor/positive pair distances. In fact, we believe having such a mix of pairs is important to creating embeddings that emphasize permutation-based relationships and not ones of distance. This approach constituted a naive approach to triplet selection -- ie, we did not worry about \textit{triplet mining} \cite{schroff_facenet_2015}. We understand that this approach may be a luxury due to the relative simplicity of our data. 

% - new subsection - - - - - - - - - - - - - - - - - - - - - - - - - - - - - - - - - - -
\subsection{Analysis}
\label{sec:approach::subsec:analysis}
We evaluated the tendency of permutation point set embeddings to cluster in two ways. The first was through a comparison of two distances, $d_{max, i}$ and $D_{ij}$, described in Figure \ref{fig:intra_inter_distances_figure}. For the comparison, we calculated the empirical cumulative distribution function\footnote{We used the CDF because it is less sensitive to choice of bin size than the probability distribution function.} (CDF) of these distances for embeddings of the permutation point sets generated by each of the four autoencoders. If the CDF of $d_{max, i}$ substantially overlaps or is greater than the CDF of $D_{ij}$, there is little to no separation of permutation point sets and no clustering on this basis is observable. However, if the $CDF(d_{max, i})$ is less than $CDF(D_{ij})$ (or mostly so), then clustering is significant and obvious. To calculate the CDF, we found the range of the distance values by calculating the minimum and maximum values of each distance set, rounded them to their floor and ceiling integers, respectively and divided that range into 1000 bins. We then counted the number of points in each distance bin and normalized the counts to the total number of points. We assumed no particular distribution form of the data in calculating the CDFs. However, the generally smooth sigmoidal shapes of most of the CDFs suggest a normal distribution of the values, though possibly skewed.

To compare $D_{ij}$ and $d_{max, i}$ from their CDF data, we calculated a 95\% centroid separation ratio given by 
\begin{equation}
R_{95\%} \;\; = \;\; \frac{D_{ij}\textit{@CDF=0.05}}{2 \; (d_{max})}
\label{eqn:centroid_separation_ratio}
\end{equation}
where $D_{ij}\textit{@CDF=0.05}$ is the value of $D_{ij}$ at a CDF value of 0.05 (or, 5\%, meaning that 95\% of the centroids are separated by a distance greater than this $D_{ij}$) and $d_{max}$ is the maximum value of $d_{max, i}$ for all $i$, which occurs at $CDF = 1.0$. If $R_{95\%} > 1.0$, there is discernible clustering. If $R_{95\%} < 1.0$, no clustering is discernible.

The degree of clustering was also visualized through plots of t-distributed stochastic neighbor embedding \cite{vanDerMaaten2008} (t-SNE) in $\mathbb{R}^{2}$ of permutation point sets and their centroids in both the original ($\mathbb{R}^{24}$) space and the embedding ($\mathbb{R}^{8}$) space. We used a \textit{perplexity} value of 150, chosen based on the rule of thumb of the square root of the number of points \cite{wattenberg2016how} ($\sqrt{25,000} \approx 150$). The number of iterations for the t-SNE calculations was 2000.

% = new section = = = = = = = = = = = = = = = = = = = = = = = = = = = = = = = = = = = = =
\section{Results and Discussion}
\label{sec:results}
% - new subsection - - - - - - - - - - - - - - - - - - - - - - - - - - - - - - - - - - -
\subsection{Effect of Introducing Triplet Loss}
\label{sec:results::subsec:add_triplet_loss}
\paragraph{On Model Accuracy.}
Adding a loss function to the model represents an additional constraint that must be accommodated while training the neural network. We expect that the added constraint would cause greater error in predictions made by the model. In Tables \ref{tab:avg_mse_loss} - \ref{tab:avg_numeric_accuracy}, we present results extracted from the model training plots above. In these tables, the averages and standard deviations were calculated from the curves above using all values over the training step interval $[600, 1000]$. Using the standard model as the baseline, we see that adding triplet loss \textit{increases the mse loss} for the enhanced models. For $\alpha = 1.0$ and $2.0$, the mse loss increases by 3\% and 5\%, respectively. At the large value of $\alpha = 5.0$, the mse loss increases superlinearly by 25\%. Clearly, the addition of the triplet loss constraint to the model increases the error in accurately reconstructing the input as the output. However, at lower values of $\alpha$, the cost may not be too high.

In the enhanced models, the fraction of the total loss that is due to the triplet loss also grows superlinearly with the value of $\alpha$. As we see in Table \ref{tab:avg_triplet_loss}, when $\alpha = 1.0$ the triplet loss is on the order of 10\% of the total loss and when $\alpha = 2.0$, it is about 17\% of the total. When $\alpha = 5.0$, the triplet loss contributes 45\% to the total loss -- almost as much as the mse loss. Remember that we used equal weighting of mse loss and triplet loss for all enhanced models. Again, the effects of increasing triplet loss on model accuracy are complicated because a higher value of $\alpha$, while introducing more error during training, is also a stronger constraint. Depending on the details of the problem, satisfying a much stronger constraint poorly might still yield results that are better, or at least not much worse, than satisfying a much weaker constraint well. Since the total loss drives the weight minimization during training, the influence of the triplet loss on adjusting the neural network weights becomes much greater as the value of $\alpha$ increases. 

\begin{table}
% table caption appears above the table
\centering
%\hrule
\caption{Average and standard deviation of steady-state minimum \textit{mse loss} during training for each model and the ratio for each enhanced model over the standard one.}
\label{tab:avg_mse_loss}
\begin{tabular}{c c c c c}
  \hline \\
 & \textbf{\textit{standard}} && \textbf{\textit{enhanced model}}  \\
 & \textbf{\textit{model}} & $\alpha$=1.0 & $\alpha$=2.0 & $\alpha$=5.0 \\ 
  \hline \\
  \textbf{average} & 0.059050 & 0.060797 & 0.062268 & 0.073508 \\
  \textbf{std dev} & 0.000009 & 0.000015 & 0.000024 & 0.000089  \\
  \multicolumn{2}{c}{\textbf{enhanced/standard ratio}} & 1.03 & 1.05 & 1.25 \\
  \hline 
\end{tabular}
%\vspace{-1em}
\end{table}
\begin{table}
% table caption appears above the table
\centering
%\hrule
\caption{Average and standard deviation of steady-state minimum \textit{triplet loss} during training and its ratio to the mse loss for each enhanced model. The ratio is calculated using the values of mse loss from Table \protect\ref{tab:avg_mse_loss}.}
\label{tab:avg_triplet_loss}
\begin{tabular}{c c c c}
  \hline \\
  && \textbf{\textit{enhanced model}}  \\
  & $\alpha$=1.0 & $\alpha$=2.0 & $\alpha$=5.0 \\ 
  \hline \\
  \textbf{average} & 0.006800	&	0.013091	&	0.059996 \\
  \textbf{std dev} & 0.000040	&	0.000053	&	0.000650 \\
  \textbf{triplet/mse ratio} & 0.112 & 0.210 & 0.816 \\
  \hline 
\end{tabular}
%\vspace{-1em}
\end{table}
\begin{table}
% table caption appears above the table
\centering
%\hrule
\caption{Average and standard deviation of steady-state minimum \textit{numerical accuracy} during training for each model and the ratio for each enhanced model over the standard one.}
\label{tab:avg_numeric_accuracy}
\begin{tabular}{c c c c c}
  \hline \\
 & \textbf{\textit{standard}} && \textbf{\textit{enhanced model}}  \\
 & \textbf{\textit{model}} & $\alpha$=1.0 & $\alpha$=2.0 & $\alpha$=5.0 \\ 
  \hline \\
  \textbf{average} & 0.825659	&	0.821290	&	0.816965	&	0.783941  \\
  \textbf{std dev} & 0.000035	&	0.000043	&	0.000069	& 0.000260   \\
  \multicolumn{2}{c}{\textbf{enhanced/standard ratio}} & 0.995 & 0.989 & 0.949  \\
  \hline 
\end{tabular}
\end{table}

\begin{figure}[htb!]
  \includegraphics[width=\textwidth]{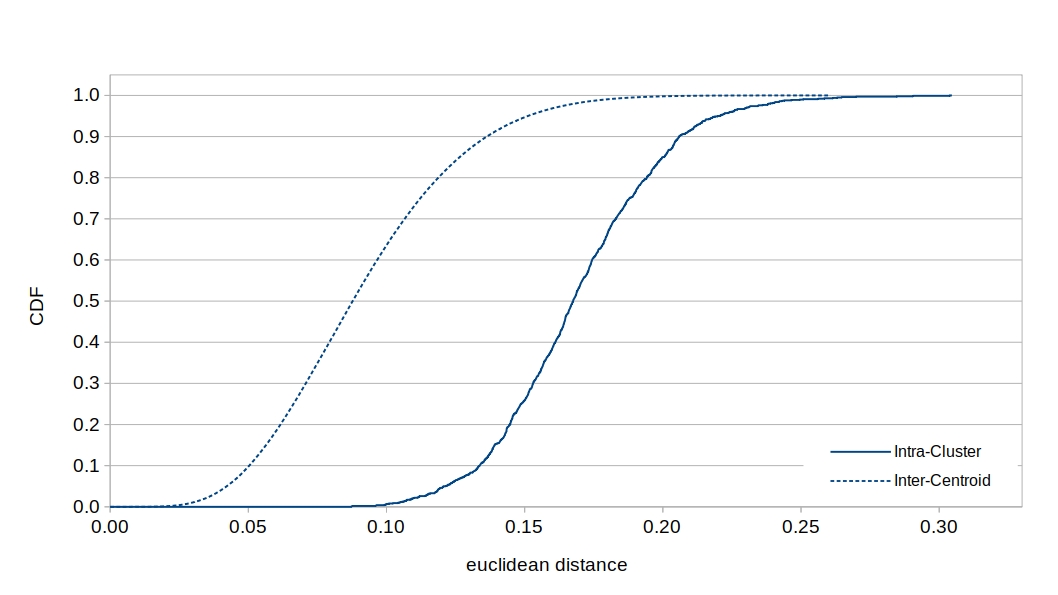}
%  \vspace*{-3em}
  \caption{Cumulative distribution functions (CDFs) of separation distances for embeddings generated by the standard autoencoder. Note that as in Figure \protect\ref{fig:intra_and_inter_distance_for_vectors_and_centroids}, the separation of point set centroids (dashed curve) is less than or in the same range as the maximum separation of a point in the permutation set from its centroid (solid curve), indicating no observable clustering.}
  \label{fig:std_loss_embedding_distance_cdf}
\end{figure}

\begin{figure}[htb!]
  \includegraphics[width=\textwidth]{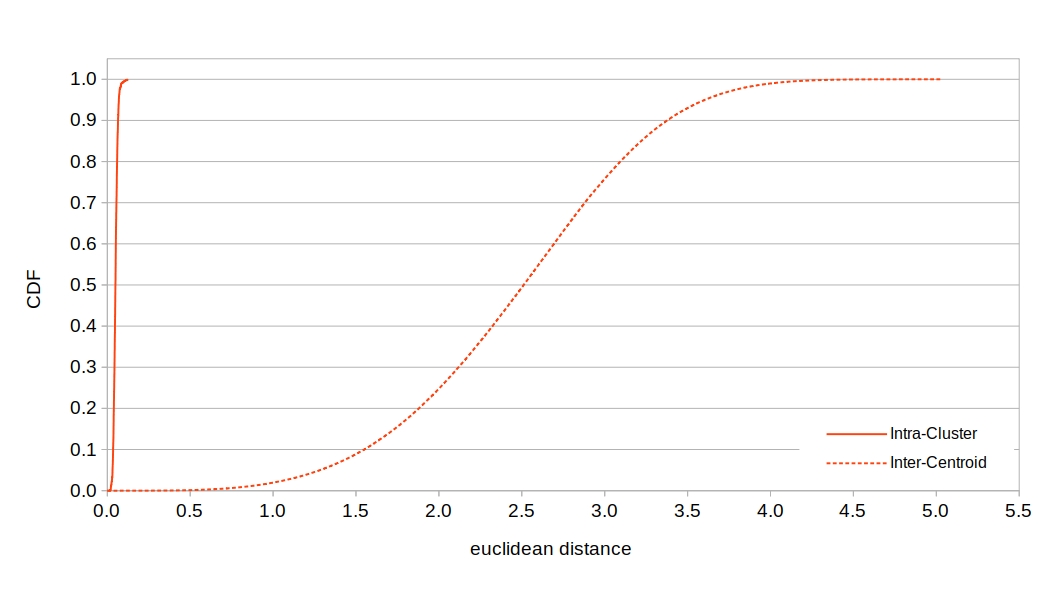}
%  \vspace*{-3em}
  \caption{Cumulative distribution functions (CDFs) of separation distances for embeddings generated by the enhanced autoencoder with triplet margin $\alpha = 1.0$. Note that in contrast to Figure \protect\ref{fig:std_loss_embedding_distance_cdf}, the maximum separation of a point in the permutation set from its centroid (solid curve) is now \textit{significantly less than} the separation of point set centroids (dashed curve). This relationship indicates significant and clearly observable clustering.}
  \label{fig:triplet_loss_alpha_1_embedding_distance_cdf}
\end{figure}

\paragraph{On Embedding Structure and Clustering.}
In Figures \ref{fig:std_loss_embedding_distance_cdf} and \ref{fig:triplet_loss_alpha_1_embedding_distance_cdf}, we show the effects of using either the standard model or enhanced models on clustering of the permutation point set embeddings. In Figure \ref{fig:std_loss_embedding_distance_cdf}, we see that the clustering behavior of the embeddings generated by the standard model are much like the original data -- there is none. Just as for the original data, $D_{ij}$ is less than or overlaps with $d_{max, i}$, indicating no discernible clustering. This result is not unexpected. The standard model uses a distance-based loss function metric (the mse loss) and we have shown that the points of one permutation point set are not expected to be nearer to each other in space than points from other permutation point sets. 

\begin{figure}[htb!] 
  \includegraphics[width=\textwidth]{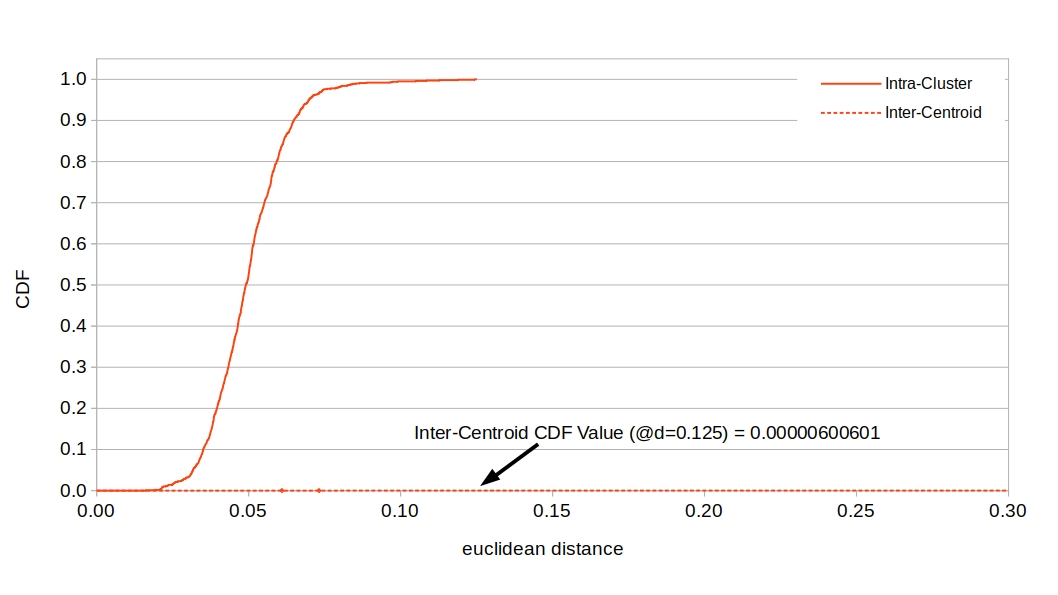}
%  \vspace*{-3em}
  \caption{Detail of Figure \protect\ref{fig:triplet_loss_alpha_1_embedding_distance_cdf}. Note that at a distance of 0.125, the value of the CDF for the intra-point set maximum separation from its centroid is 1.0, meaning 100\% of all embedding $d_{max, i}$ are less than 0.125, while the CDF value for the inter-centroid distances $\approx$  $6 \times 10^{-6}$, meaning less than 0.0006\% of all embedding $D_{ij}$ have a separation of 0.125. Given that there are 24,000 embedding points, there are only one or two at this distance range. The vast majority of embedding space centroids have a separation of much more than 0.125. Clearly, there is significant clustering by permutation point set for these embeddings.}
  \label{fig:triplet_loss_alpha_1_embedding_distance_cdf-2}
\end{figure}

However with the addition of triplet loss where positive points belong to the same permutation point set as the anchor points and negative points belong to different permutation point sets, we have introduced a constraint based on the subvector permutation relationship (even if this constraint is expressed as a type of distance metric as in Equation \ref{eqn:triplet_loss}). For the enhanced models trained with triplet loss margin $\alpha = 1.0$, (Figure \ref{fig:triplet_loss_alpha_1_embedding_distance_cdf}) the results are dramatic. The relationship between $D_{ij}$ and $d_{max,i}$ flips, and strong clustering of the embeddings based on subvector permutation appears. In fact, the clusters are so well separated that there is only on embedding point from one cluster that overlaps with one other cluster out of 1000 sets of points (24,000 points total), as indicated in Figure \ref{fig:triplet_loss_alpha_1_embedding_distance_cdf-2}.


\begin{figure}[t] 
  \includegraphics[width=\textwidth]{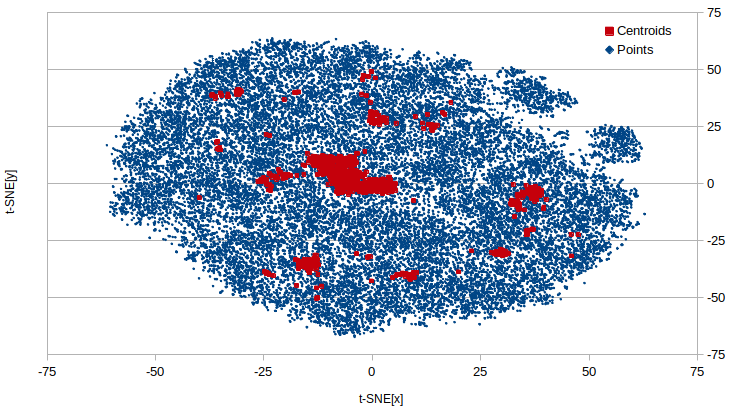}
  \caption{Visualization of the $\mathbb{R}^{8}$ embedding space using t-distributed stochastic neighbor embedding (t-SNE) in $\mathbb{R}^{2}$ for the standard autoencoder. The red points are the centroids of the embedding permutation point sets. The blue points are the embeddings of all 1000 vectors and their permutations (24,000 points total). Clearly, no permutation set-based clustering of any kind is evident in this set of embeddings, consistent with evidence discussed above.}
  \label{fig:std_loss_embedding_tsne}
\end{figure}

\begin{figure}[htb!] 
  \includegraphics[width=\textwidth]{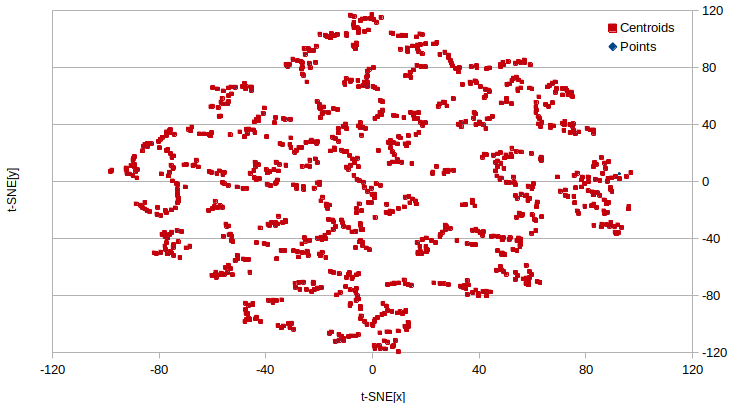}
  \caption{Visualization of the $\mathbb{R}^{8}$ embedding space using t-distributed stochastic neighbor embedding (t-SNE) in $\mathbb{R}^{2}$ for the enhanced autoencoder with triplet loss margin $\alpha$ = 1.0. The red points are the centroids of the embedding permutation point sets. The blue points are the embeddings of all 1000 vectors and their permutations (24,000 points total) and are almost completely hidden behind their centroids. Clearly, there is significant clustering by permutation point set for these embeddings, consistent with evidence discussed above.}
  \label{fig:triplet_loss_alpha_1_embedding_tsne}
\end{figure}

We can also see the dramatic clustering using t-SNE visualization in Figures \ref{fig:std_loss_embedding_tsne} and \ref{fig:triplet_loss_alpha_1_embedding_tsne}. In both figures, the blue points are the embedding values and the red points are the centroids of the embedding permutation point sets. In Figure \ref{fig:std_loss_embedding_tsne}, we see that for the standard model, there is no discernible clustering of the embeddings -- there is more or less a uniform sea of blue points. We observe some coarse clustering of the centroids, but the maximum diameters ($d_{max, i}$) of the permutation point set embeddings are so much larger than the size of the centroid clusters it renders any centroid clustering irrelevant and uninteresting. However, when triplet loss is added in the enhanced model, the embedding clustering around permutation point sets is profound, as seen in Figure \ref{fig:triplet_loss_alpha_1_embedding_tsne}. In this plot, the blue points are not omitted -- they are almost completely hidden behind the red points -- their cluster centroids! This plot reinforces the results from Figure \ref{fig:triplet_loss_alpha_1_embedding_distance_cdf} that the flip of $D_{ij} <= d_{max, i}$ in the original data and standard model embeddings to $D_{ij} >> d_{max, i}$ for the enhanced model with $\alpha = 1.0$.


% - new subsection - - - - - - - - - - - - - - - - - - - - - - - - - - - - - - - - - - -
\subsection{Effect of Triplet Loss Margin $\alpha$}
\label{sec:results::subsec:triplet_alpha}
\begin{figure}[htb!] 
  \includegraphics[width=\textwidth]{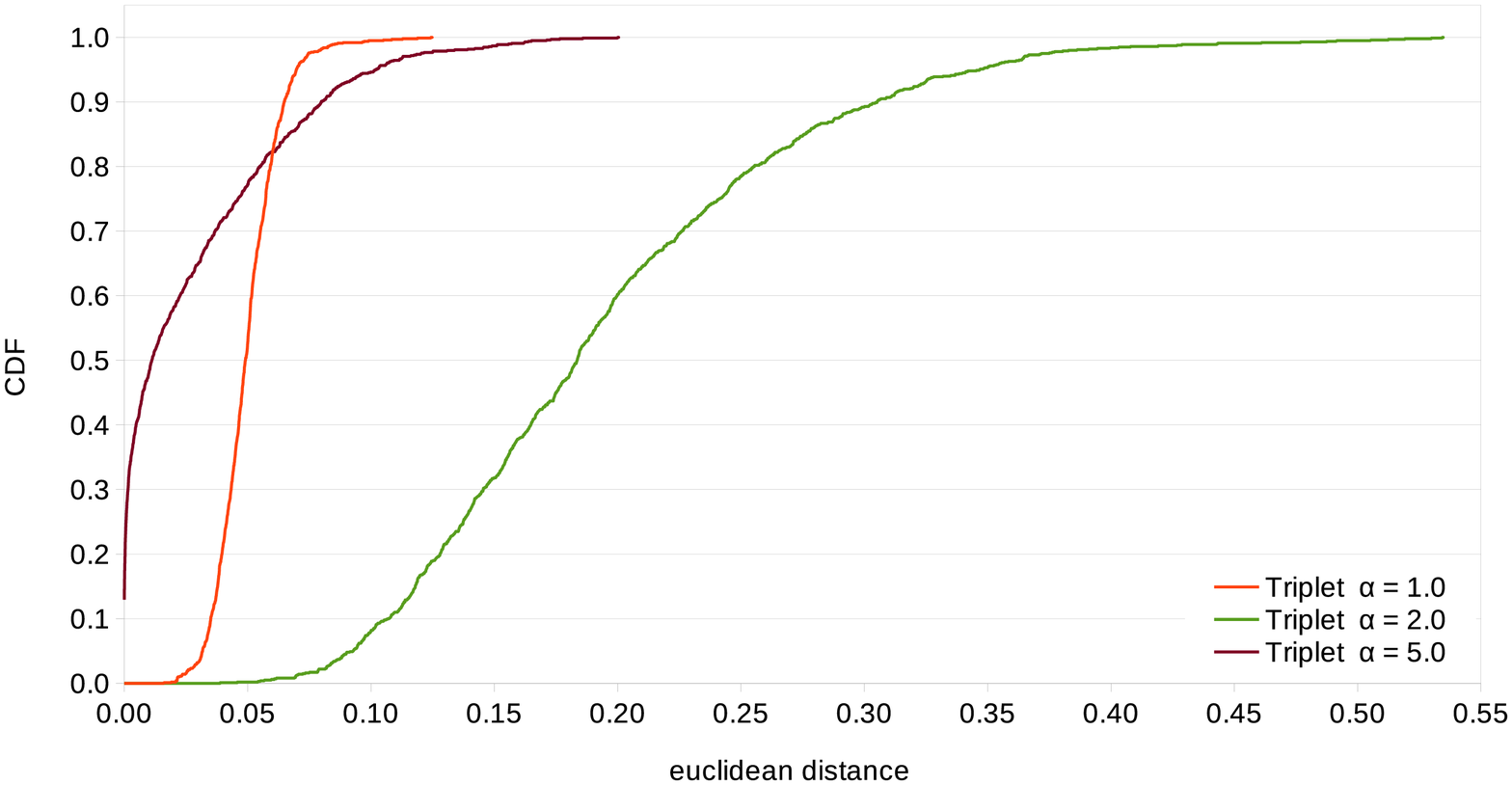}
  \caption{Comparison of the cumulative distribution functions for the intra-cluster maximum separation $d_{max, i}$ of permutation point set embeddings for values of triplet loss margin $\alpha$ of \{1.0, 2.0, 5.0\}.}
  \label{fig:intra_cluster_max_triplet_embedding}
\end{figure}

The value of the triplet loss margin $\alpha$ had a clear but not necessarily simple effect on the embeddings. Figure \ref{fig:intra_cluster_max_triplet_embedding} shows the effect of $\alpha$ on the maximum size of permutation point set embedding clusters, $d_{max, i}$. For $\alpha = 1.0$, the median value of $d_{max, i}$ (as given by $d_{max, i}$ @CDF = 0.5) is small -- about $0.05$. The CDF is mostly vertically symmetric around CDF = 0.5 and ranges over a short distance (ie, the difference between $d_{max, i}$ @CDF = 0 and $d_{max, i}$ @CDF = 1) of about $0.1$, indicating tight clustering around the centroids, as seen in Figure \ref{fig:triplet_loss_alpha_1_embedding_tsne}. When $\alpha = 2.0$, the clusters become larger, with a median $d_{max, i}$ of about $0.18$, and the CDF develops a long tail given by its vertical \textit{asymmetry}, indicating that cluster size \textit{increases faster} when above the median value than below it. However, when $\alpha = 5.0$, the median $d_{max, i}$ shrinks drastically to about $0.01$ -- much smaller even than for $\alpha = 1.0$, and it exhibits very strong long tail. Thus, as $\alpha$ increases over this range, the effect on embedding cluster sizes varies widely and not monotonically. They start small and tight, become slightly larger and more dispersed, then become very small and (relatively speaking) very dispersed. A more detailed investigation of these effects will be the subject of future work.

\begin{figure}[htb!] 
  \includegraphics[width=\textwidth]{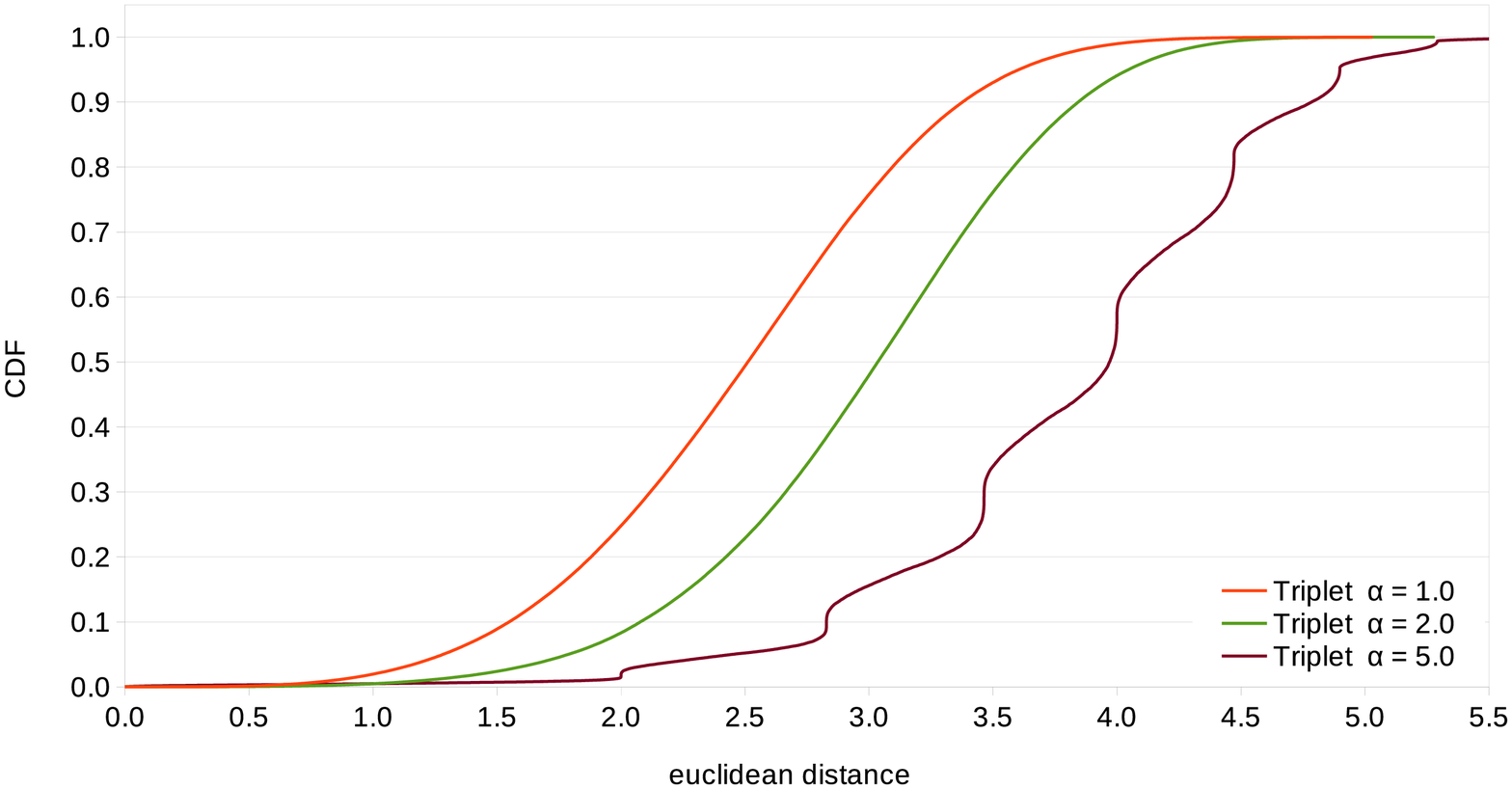}
  \caption{Comparison of the cumulative distribution functions for the inter-cluster centroid separation $D_{ij}$  of permutation point set embeddings for values of triplet loss margin $\alpha$ of \{1.0, 2.0, 5.0\}.}
  \label{fig:inter_centroid_triplet_embedding}
\end{figure}

Likewise, the value of the triplet loss margin $\alpha$ has a strong effect on $D_{ij}$, the separation of permutation point set embedding centroids. As shown in Figure \ref{fig:inter_centroid_triplet_embedding}, the average centroid separation increases as $\alpha$ increases from 1.0 to 2.0. However, the CDF for $\alpha$ = 5.0 exhibits a smoothed stair-step characteristic that reflects additional structure in the clustering behavior beyond that observed for the other values of $\alpha$. For this curve, we observe a series of smooth rises followed by sharp increases starting at $D_{ij} \approx 2.0$. This behavior indicates a multimodal distribution with alternating sharp and broad peaks in the distribution corresponding to successive tight and dispersed characteristic clustering distances. For example, at $D_{ij} \approx 2.8$, there is a sharp, nearly vertical rise in the CDF from $\approx$ 0.09 to $\approx$ 0.11, a difference of $\approx$ 0.02. Thus, about 2\% of centroid pairs have a separation distance of $\approx$ 2.8 within a narrow error range, given by the sharpness of the rise. The gradual rise from $D_{ij} \approx 2.8$ to $D_{ij} \approx 3.6$ indicates a broader peak centered at a distance of $\approx$ 3.2 that is $\approx$ 15\% of the total. We do not yet know if this structure is spurious or real. It may well be due to real effects, such as a superclustering of separate permutation point sets that have some values in common or all values in common, but not in the correct order to be related by subvector permutation. Or, it may completely be an artifact of the relatively large value of $\alpha$. This effect will be the topic of future investigation.

\begin{figure}[htb!] 
  \includegraphics[width=\textwidth]{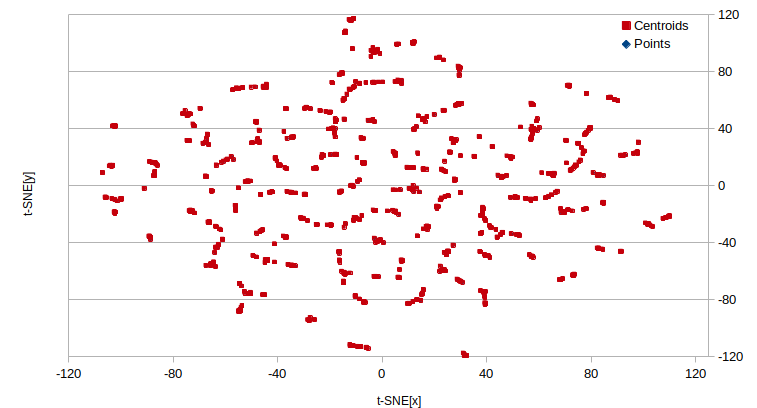}
  \caption{Visualization of the embedding space using t-distributed stochastic neighbor embedding (t-SNE) in $\mathbb{R}^{2}$ for the enhanced autoencoder with triplet loss margin $\alpha$ = 5.0. The red points are the centroids of the embedding permutation point sets. The blue points are the embeddings of all 1000 vectors and their permutations (24,000 points total) and are completely hidden behind their centroids. There is significant clustering by permutation point set for these embeddings, as well as clustering of the centroids, leading to the step-like property of the CDF in Figure \protect\ref{fig:inter_centroid_triplet_embedding}.}
  \label{fig:triplet_loss_alpha_5_embedding_tsne}
\end{figure}

Turning to the t-SNE visualization in Figure \ref{fig:triplet_loss_alpha_5_embedding_tsne}, we observe a superclustering of centroids in addition to their permutation point set embeddings, ranging from two to several centroids (and so permutation point set embedding clusters) per supercluster. However, the permutation point set embedding clusters are still distinct as evidenced above by $D_{ij} >> d_{max, i}$ for $\alpha$ = 5.0. It is this superclustering that gives rise to the step-like behavior of the CDF above. 

\paragraph{The 95\% Centroid Separation Ratio.}
The 95\% centroid separation ratio, $R_{95\%}$, given in Equation \ref{eqn:centroid_separation_ratio} indicates the extent of clustering. For $R_{95\%} < 1$, there is significant overlap of points in different clusters, at $R_{95\%} = 1$, the clusters are just separable, and at $R_{95\%} > 1$, the clusters are well separated. For both the original data and the embeddings from the standard model, $R_{95\%}$ is less than 1 ($\approx$ 0.15 and $\approx$ 0.25 respectively. For the embeddings from the enhanced models, $R_{95\%}$ is greater than 1, but, as seen above, the results are complicated. For $\alpha = 1.0$, $2.0$, and $5.0$, the values of $R_{95\%}$ are approximately 5.0, 1.6 and 6.6, respectively, showing strong to very strong separation of the permutation-based embedding clusters.

% = new section = = = = = = = = = = = = = = = = = = = = = = = = = = = = = = = = = = = = =
\section{Conclusions}
\label{sec:conclusions}
In this work, we have shown that an autoencoder enhanced with triplet loss is an effective means to generate invariant representations of vectors that are related by subvector permutation, where a standard autoencoder fails to capture this pattern. In the original data and the embeddings produced by the standard autoencoder, no clustering of data points was discernible. The invariance was achieved by the triplet loss driving the permutation-related vector embeddings to tightly cluster around a single value -- the centroid -- with relatively large separation of centroids relative to the maximum value of the cluster radii. At the lowest value of $\alpha$ studied ($\alpha = 1.0$), clustering of the embeddings was significant. This result is somewhat surprising, because, in the standard model, the distance-based constraint (mse loss) had no effect on permutation-based clustering, but adding a second distance-based constraint (triplet loss), where the distances are based on permutation point set relationships, had a dramatic effect. The 95\% centroid separation ratio ($R_{95\%}$) with even this weak constraint was $R_{95\%} \approx 5$, where the minimum value for clear separation is $1.0$. 

The value of  $\alpha$ used in the enhanced autoencoder model had a significant and complicated affect on the embedding clustering. 
For $\alpha = 1.0$, the clusters were small and tight, with a maximum radius of about 0.125 and average cluster separation of 2.5. When $\alpha = 2.0$, the clusters were larger in size and spaced farther apart, but still clearly separated. However, when $\alpha = 5.0$, the clusters were significantly smaller than even for $\alpha = 1.0$ and spaced much farther apart. There was clear asymmetry in the maximum cluster radius distribution. The clusters themselves exhibited superclustering behavior as evidenced in the stair-step characteristic of the cluster separation distribution. This behavior may or may not be an artifact of the relatively large value of $\alpha$ and warrants further investigation. 

Finally, for our data set and with triplet loss margin $\alpha = 1.0$, $\epsilon \approx 0.125$, the maximum size of the permutation point set embedding clusters. Thus, two vectors from the original data set that are permutations of each other as described above will map to the same embedding value to within an error of $\epsilon$, making the embedding value an invariant of subvector permutation. This error may be reduced for a better tuned autoencoder (see \textit{Future Work} below). Regardless, it is an effective means of identifying clusters of vector permutations where the original vectors are randomly generated.

\section{Future Work}
\label{sec:future_work}
We are currently using enhanced autoencoder described here (with triplet loss margin $\alpha = 1.0$) to generate embeddings of single turn game vectors in our original application. Sequences of these embeddings will then be used to generate embeddings of game turn sequences using an LSTM-based autoencoder, since games may be of variable length. However, as discussed above, the value chosen for the the triplet loss margin $\alpha$ causes significant and varied effects on the characteristics of the embeddings. Better understanding the origin and nature of some of these effects is a subject of future investigation, including applying smaller values of $\alpha$ that were investigated here. For example, we might compare the (equal) static weighting of mse loss and triplet loss as we have used here with a more dynamic weighting to keep the triplet loss at a constant (low) fraction of the total loss, independent of value of $\alpha$. Would such an approach mitigate the distortions at high value of $\alpha$?

We have also started examining whether adding a second stage of training for enhanced autoendocer using \textit{centroid values} as output is useful to produce an even more accurate invariant representation. If \textit{all} permutation point set embeddings cluster around central values (ie, their centroids), then those values become \textit{de facto} invariant representations. Again, any second stage of training of an autoencoder that does not use original values as output\footnote{Technically, the model would not be an \textit{auto}\hspace{1pt}encoder if the output is not the same as the input.} will cause a lower predictive accuracy of the model. The key questions are how much less accurate and whether that price is too high, meaning that the predictions using the ``tuned'' embedding values are unacceptably less accurate. 

A related issue is the effect of noise on the ability to generate invariant embeddings. In the data set we used here, there was no noise in values of the vector elements of the data set in this study because that is the true of the game turn data set it is meant to model. However, to be generally applicable, the enhanced autoencoder generating subvector permutation invariant embeddings should be resistant to noisy data. We plan to run further experiments by adding various levels of noise to the training/testing data and validation data.

%\newpage
% = new section = = = = = = = = = = = = = = = = = = = = = = = = = = = = = = = = = = = = =

% Authors must disclose all relationships or interests that 
% could have direct or potential influence or impart bias on 
% the work: 
%
\section*{Funding}
This work was funded in part through the JHU APL Project Catalyst Propulsion Grant program.

\section*{Conflict of Interest}
The authors declare that they have no conflicts of interest regarding the content of this paper.

\section*{Acknowledgments}
The author thanks JHU APL for partial financial support of this work through the Project Catalyst Propulsion Grant program and Will Ames of JHU APL for securing funding for his Propulsion Grant proposal and allowing me to work on it. The author also thanks Will Ames, Kyle Casterline, Jason Scott, John Winder, and Tom Urban of JHU APL for reviewing this manuscript and providing valuable feedback. Finally, the author thanks Kyle Casterline and Will Li for bringing the triplet loss method to the author's attention.

\bibliographystyle{unsrt}  
\bibliography{references}

\begin{thebibliography}{10}

\bibitem{Goodfellow2016Chapter14}
Ian Goodfellow, Yoshua Bengio, and Aaron Courville.
\newblock {\em Deep Learning}.
\newblock MIT Press, 2016.

\bibitem{TschannenAutoencoderAdvances_Proceedings}
Michael Tschannen, Olivier~Frederic Bachem, and Mario Lučić.
\newblock Recent advances in autoencoder-based representation learning.
\newblock In {\em Bayesian Deep Learning Workshop, NeurIPS}, 2018.

\bibitem{chechik2010large}
Gal Chechik, Varun Sharma, Uri Shalit, and Samy Bengio.
\newblock Large scale online learning of image similarity through ranking.
\newblock {\em Journal of Machine Learning Research}, 11:1109--1135, 2010.

\bibitem{schroff_facenet_2015}
Florian Schroff, Dmitry Kalenichenko, and James Philbin.
\newblock {FaceNet}: {A} {Unified} {Embedding} for {Face} {Recognition} and
  {Clustering}.
\newblock {\em 2015 IEEE Conference on Computer Vision and Pattern Recognition
  (CVPR)}, pages 815--823, June 2015.
\newblock arXiv: 1503.03832.

\bibitem{hoffer_deep_2015_proceedings}
Elad Hoffer and Nir Ailon.
\newblock Deep metric learning using triplet network.
\newblock In Aasa Feragen, Marcello Pelillo, and Marco Loog, editors, {\em
  Similarity-Based Pattern Recognition}, pages 84--92, Cham, 2015. Springer
  International Publishing.

\bibitem{ishfaq_deep_nodate}
Haque Ishfaq and Ruishan Liu.
\newblock Deep {Metric} {Learning} with {Triplet} {Loss} and {Variational}
  {Autoencoder}.

\bibitem{kaya_deep_2019}
{Kaya} and {Bilge}.
\newblock Deep {Metric} {Learning}: {A} {Survey}.
\newblock {\em Symmetry}, 11(9):1066, August 2019.

\bibitem{KumarSemiSupervisedClusteringWithMetricLearning2005}
N.~Kumar, Krishna Kummamuru, and D.~Paranjpe.
\newblock Semi-supervised clustering with metric learning using relative
  comparisons.
\newblock page 4 pp., 12 2005.

\bibitem{dong2018triplet}
Xingping Dong and Jianbing Shen.
\newblock Triplet loss in siamese network for object tracking.
\newblock In {\em Proceedings of the European Conference on Computer Vision
  (ECCV)}, pages 459--474, 2018.

\bibitem{hermans_defense_2017}
Alexander Hermans, Lucas Beyer, and Bastian Leibe.
\newblock In {Defense} of the {Triplet} {Loss} for {Person}
  {Re}-{Identification}.
\newblock {\em arXiv:1703.07737 [cs]}, November 2017.
\newblock arXiv: 1703.07737.

\bibitem{yang_triplet_2019}
Yao Yang, Haoran Chen, and Junming Shao.
\newblock Triplet {Enhanced} {AutoEncoder}: {Model}-free {Discriminative}
  {Network} {Embedding}.
\newblock In {\em Proceedings of the {Twenty}-{Eighth} {International} {Joint}
  {Conference} on {Artificial} {Intelligence}}, pages 5363--5369, Macao, China,
  August 2019. International Joint Conferences on Artificial Intelligence
  Organization.

\bibitem{ienco_deep_triplet_2019}
Dino Ienco and Ruggero~G. Pensa.
\newblock Deep {Triplet}-{Driven} {Semi}-supervised {Embedding} {Clustering}.
\newblock In Petra Kralj~Novak, Tomislav Šmuc, and Sašo Džeroski, editors,
  {\em Discovery {Science}}, volume 11828, pages 220--234. Springer
  International Publishing, Cham, 2019.
\newblock Series Title: Lecture Notes in Computer Science.

\bibitem{python3}
Guido Van~Rossum and Fred~L. Drake.
\newblock {\em Python 3 Reference Manual}.
\newblock CreateSpace, Scotts Valley, CA, 2009.

\bibitem{chollet2015keras}
Fran\c{c}ois Chollet et~al.
\newblock Keras.
\newblock \url{https://keras.io}, 2015.

\bibitem{tensorflow2015-whitepaper}
Mart\'{\i}n Abadi, Ashish Agarwal, Paul Barham, Eugene Brevdo, Zhifeng Chen,
  Craig Citro, Greg~S. Corrado, Andy Davis, Jeffrey Dean, Matthieu Devin,
  Sanjay Ghemawat, Ian Goodfellow, Andrew Harp, Geoffrey Irving, Michael Isard,
  Yangqing Jia, Rafal Jozefowicz, Lukasz Kaiser, Manjunath Kudlur, Josh
  Levenberg, Dandelion Man\'{e}, Rajat Monga, Sherry Moore, Derek Murray, Chris
  Olah, Mike Schuster, Jonathon Shlens, Benoit Steiner, Ilya Sutskever, Kunal
  Talwar, Paul Tucker, Vincent Vanhoucke, Vijay Vasudevan, Fernanda Vi\'{e}gas,
  Oriol Vinyals, Pete Warden, Martin Wattenberg, Martin Wicke, Yuan Yu, and
  Xiaoqiang Zheng.
\newblock {TensorFlow}: Large-scale machine learning on heterogeneous systems,
  2015.
\newblock Software available from tensorflow.org.

\bibitem{numpy}
Travis Oliphant.
\newblock {NumPy}: A guide to {NumPy}.
\newblock USA: Trelgol Publishing, 2006--.
\newblock [Online; accessed 17 May 2020].

\bibitem{scikit-learn}
F.~Pedregosa, G.~Varoquaux, A.~Gramfort, V.~Michel, B.~Thirion, O.~Grisel,
  M.~Blondel, P.~Prettenhofer, R.~Weiss, V.~Dubourg, J.~Vanderplas, A.~Passos,
  D.~Cournapeau, M.~Brucher, M.~Perrot, and E.~Duchesnay.
\newblock Scikit-learn: Machine learning in {P}ython.
\newblock {\em Journal of Machine Learning Research}, 12:2825--2830, 2011.

\bibitem{vanDerMaaten2008}
Laurens van~der Maaten and Geoffrey Hinton.
\newblock Visualizing data using {t-SNE}.
\newblock {\em Journal of Machine Learning Research}, 9:2579--2605, 2008.

\bibitem{wattenberg2016how}
Martin Wattenberg, Fernanda Viégas, and Ian Johnson.
\newblock How to use t-sne effectively.
\newblock {\em Distill}, 2016.

\end{thebibliography}

\end{document}